\pdfoutput=1
\documentclass[11pt]{article}
\usepackage{arabtex}
\usepackage{utf8}

\usepackage{emnlp2021}
\setcode{utf8}

\usepackage{times}
\usepackage{latexsym}
\usepackage{amsmath}
\usepackage{subcaption}
\usepackage{graphicx}
\usepackage{multirow}
\usepackage[T1]{fontenc}
\usepackage{url}
\usepackage[inline]{enumitem}
\usepackage{relsize}
\usepackage[utf8]{inputenc}
\usepackage{microtype}

\usepackage{marginnote}
\usepackage{xcolor}

\title{Using Optimal Transport as Alignment Objective for fine-tuning Multilingual Contextualized Embeddings}

\author{Sawsan Alqahtani, Garima Lalwani, Yi Zhang, Salvatore Romeo, Saab Mansour\\
  AWS, Amazon AI \\
  \texttt{{sawsa,glalwani,yizhngn,romeosr,saabm}@amazon.com} \\}

\begin{document}
\maketitle
\begin{abstract}
Recent studies have proposed different methods to improve multilingual word representations in contextualized settings including techniques that align between source and target embedding spaces. For contextualized embeddings, alignment becomes more complex as we additionally take context into consideration. In this work, we propose using Optimal Transport  (OT) as an alignment objective during fine-tuning to further improve multilingual contextualized representations for downstream cross-lingual transfer. This approach does not require word-alignment pairs prior to fine-tuning that may lead to sub-optimal matching and instead learns the word alignments within context in an unsupervised manner. It also allows different types of mappings %(e.g. one to many, many to one) 
due to soft matching between source and target sentences. We benchmark our proposed method on two tasks (XNLI and XQuAD) and achieve improvements over baselines as well as competitive results compared to similar recent works. 
\end{abstract}

\section{Introduction}
Contextualized word embeddings have advanced the state-of-the-art performance in different NLP tasks \cite{peters2018deep,howard2018universal,devlin2019bert}. Similar advancements have been made for languages other than English using models that learn cross-lingual word representations leveraging monolingual and/or parallel data \cite{devlin2019bert, Conneau2020UnsupervisedCR, artetxe2020crosslingual}. Such cross-lingual ability helps in mitigating the lack of abundant data (labelled or  unlabelled) and computational resources for languages other than English, with lesser cost.
%The cross-lingual model ability of the model is essential to mitigate, with less cost, the lack of existing data and computational resources for languages other than English. 
Yet, there exists a challenge for  improving multilingual representations and cross-lingual transfer learning, especially for low resource languages. %remains. 
Recent studies proposed different techniques to improve multilingual representations in contextualized settings with additional objectives such as translation language modeling \cite{lample2019crosslingual}, %sentence order prediction \cite{lan2020albert},
integrating language and task adapters \cite{pfeiffer2020madx}, and applying alignment techniques in the embedding spaces \cite{cao2020multilingual,wu2020explicit}.

\begin{figure}
    \centering
  \includegraphics[width=\linewidth]{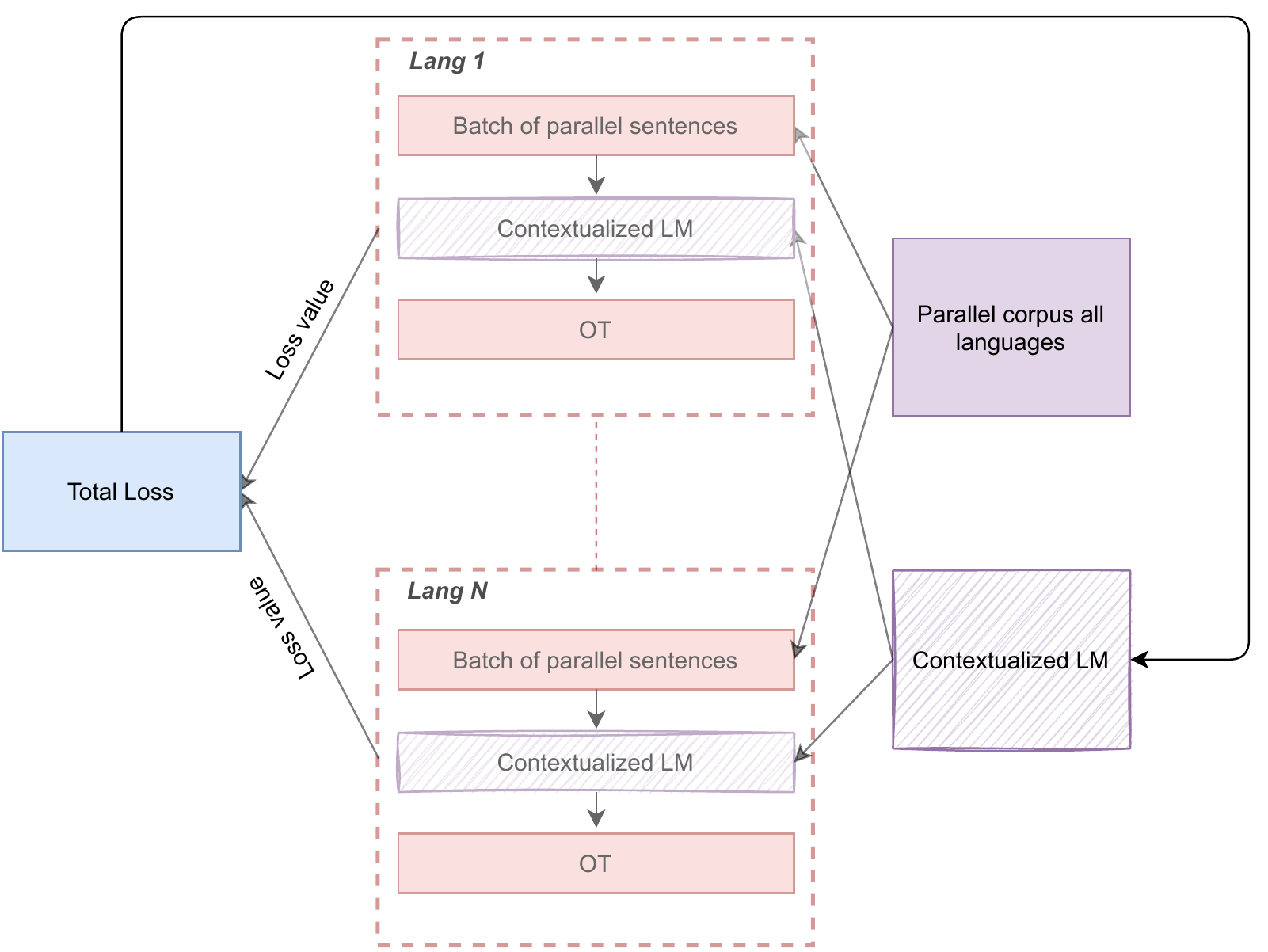}
  \caption{One iteration of fine-tuning a contextualized language model (LM) with optimal transport (OT) as the loss value.}
  \vspace{-7mm}
 % \caption{The process of fine-tuning a contextual language model with optimal transport (OT). Independent OT is learnt per language pair, resulting in loss values that reflect the cost of transferring knowledge  from source to target. In each iteration, we apply the fine-tuned model on sentence pairs to generate their contextualized embeddings for the next optimization iteration.}
  \label{finetuning_ot}
\end{figure}
Previous studies concerning alignment in the embedding space show promising directions to improve cross-lingual transfer abilities for low resource languages \cite{aldarmaki-etal-2018-unsupervised, Schuster2019CrossLingualAO, wang2019crosslingual, cao2020multilingual}. The objective is to align source and target language representations into the same embedding space, for instance by encouraging similar words to be closer to each other (e.g. \textit{cat} in English and \textit{Katze} in German) with least cost in terms of data and computational resources. Such methods require some form of cross-lingual signal, such as alignment in non-contextualized embeddings, mainly utilize  bilingual/multilingual lexicon that have been learned with unsupervised or supervised techniques   \cite{mikolov2013exploiting,smith2017offline,aldarmaki-etal-2018-unsupervised}. However, when it comes to contextualized embeddings, alignment becomes more complex as we additionally utilize context (e.g. \textit{``match''} in \textit{``Your shoes don't \underline{match} clothes''} is similar to  word \textit{``passen''} in \textit{``Ihre Schuhe \underline{passen} nicht zu Kleidung''} but not to ``match'' in \textit{``Hast du das Cricket \underline{Match} gesehen?''}). In this work,  we use only parallel sentences as an informative cross-lingual training signal.

Along these lines, previous studies mainly followed two approaches: (1) rotation based techniques with the Procrustes objective where the source embedding space is rotated to match that of the target \cite{wang2019crosslingual,Aldarmaki2019ContextAwareCM}; (2) fine-tuning the pre-trained language model (LM) with explicit alignment objectives such that similar words in parallel sentences are closer in representation space \cite{cao2020multilingual,wu2020explicit,hu2021explicit}. Fine-tuning with alignment objective function provides simple yet effective and promising solution to improve  contextualized word representations, especially for low resource languages. As opposed to rotation based approaches which require generating a transformation matrix for each language pair of interest, the alignment objective allows simultaneous learning from multiple languages. 

\begin{figure}
\centering
  \includegraphics[width=\linewidth]{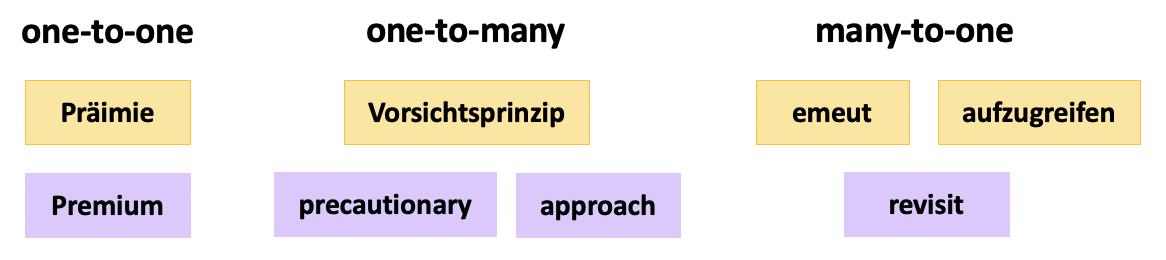}
  \caption{Examples of word alignments between English and German.}
  \vspace{-5mm}
  \label{relation_examples}
\end{figure}

Majority of previous studies concerning fine-tuning with alignment objective start with pre-collected word pairs generated using unsupervised or supervised methods (e.g. fastAlign \cite{dyer-etal-2013-simple}) which aligns words in source and target sentences based on semantics, and subsequently applies some heuristics to obtain one-to-one word alignments. However, this leads to losing other word relationships (e.g. many-to-one) which exist in some language pairs (Figure \ref{relation_examples})

Inspired by the limitations of previous works, we propose the use of optimal transport (OT henceforward) to transfer knowledge across languages and improve multilingual word representation for cross-lingual transfer in zero-shot setting. 
%GL: Removed few shot since we don't evaluate with this setting. 
This method learns word alignments while fine-tuning the pre-trained representations in an end-to-end fashion. As opposed to previous studies, this eliminates the need for pre-collected word pairs and allows many-to-many mappings between source and target words. Furthermore, our approach directly utilizes the continuous representation of contextualized word embeddings for alignment which helped broaden the scope of alignments to include additional linguistic information embedded in the LM  (e.g. semantic and syntactic structure). % by leveraging word representations from last layer of contextualized LMs. This helped in broadening the scope of alignments to include additional linguistic information embedded in the last layer  (e.g. semantic and syntactic structure). 
%GL: There was sudden jump to last layer with no mention before it as to how we use contextualized embeddings (from which layer etc.). Added this reference before
%TODO: Should we mention about your experiments that considered multiple layers? 

Specifically, we optimize a regularized variant of OT, i.e. Sinkhorn divergence \cite{feydy2019interpolating}, on parallel sentences and use that as a guidance to fine-tune the pre-trained LM. We learn several independent OT mappings per language pair, each guiding the model to further shift contextualized word embeddings in the source language towards the ones in the target language (refer to Figure \ref{finetuning_ot}). Compared to the baseline mBERT, we obtain improvements of 1.9\% and 1.3\% F1 on average in XNLI \cite{conneau2018xnli} and XQuAD \cite{rajpurkar2016squad, artetxe2020crosslingual} benchmarks, respectively.

Before we dive deep into our method (Section \ref{methods}), we briefly describe OT in Section \ref{background} and related work in Section \ref{related_work}. We discuss the experimental setup, results, analysis and finally conclusion in Sections \ref{setup}, \ref{results},  \ref{analysis} and \ref{conclusion} respectively. Our contribution is mainly three-fold:
\begin{itemize}[leftmargin=*]
    \setlength\itemsep{0.5mm}
    \item We propose the use of OT to align source and target embeddings in an unsupervised fashion eliminating the need for pre-collected one-to-one word pairs,
    \item We use OT within the space of contextual embeddings in an end-to-end manner by leveraging loss from OT optimization for fine-tuning contextualized embeddings.
    \item We show  improvements compared to the baselines and competitive results compared to more recent works evaluating on  XNLI and XQuAD. 
    %\item We provide an extensive analysis for our proposed methods.  
\end{itemize}

\section{Optimal Transport in NLP}
\label{background}
Optimal transport (OT) provides powerful tools to compare between different probability distributions %on $R^d$ 
and learn the similarities/differences to move the mass from source to target distributions \cite{peyre2020computational}. It has a strict requirement where source distribution must be completely transferred to target distribution  making it rigid for machine learning based models. \newcite{santambrogio2015optimal} relaxed this constraint by allowing masses to be partially transferred to more than one point in the target distribution resulting in development of different regularized OT variants that improve both computational and statistical properties %which makes it suitable for machine learning based models 
\cite{cuturi2013sinkhorn,schmitzer2019stabilized,alaya2020screening}.

Regularized OT variants led to the increased adoption of OT %, allowing its applicability 
in a wide range of areas such as graph applications, computer vision, and natural language processing \cite{vayer2019optimal,xu2019learning,bcigneul2021optimal,singh2020context,alvarezmelis2020unsupervised}.
Downstream applications utilize OT properties to obtain either of the following outputs \cite{flamary2021pot}: (1) optimal mapping (or transport) that aligns between points in the two distributions for applications like graph matching \cite{vayer2019optimal} or word translations \cite{alvarezmelis2018gromovwasserstein}; (2) the optimal values (Wasserstein distance) which is computed based on the optimal mappings and subsequently guide the model learning for applications like document similarity \cite{pmlr-v37-kusnerb15} and machine translation \cite{chen2019improving}. Similar to this line of work, we use optimal values for guiding model learning.

\paragraph{Optimal transport for alignment:} OT has been used as a fully unsupervised algorithm to align points between two distributions making use of the linguistic and structural similarity for applications like bilingual lexical induction  \citep{zhang-etal-2017-earth,Schmitz_2018} and word translations \citep{alvarezmelis2018gromovwasserstein,alvarezmelis2019optimal,grave2018unsupervised}. For instance, \citet{zhang-etal-2017-earth} use OT to induce translations for a source word from the target language in bilingual lexicon induction whereas \citet{xu2018unsupervised} use OT and back-translation losses to align traditional monolingual word embeddings that do not leverage context (i.e. word type level).

\section{Related Work}
\label{related_work}
Alignment as a post-processing technique on  distributional embedding spaces provides an effective solution to improve cross-lingual downstream applications. 
%Recent efforts develop models to learn cross-lingual word representations, leveraging monolingual and/or parallel data \cite{devlin2019bert, Conneau2020UnsupervisedCR, artetxe2020crosslingual, pfeiffer2020madx}. \citet{artetxe2020rigor} challenged the motivation of  unsupervised cross-lingual learning using only monolingual data suggesting that no parallel data and sufficient monolingual data for low resource languages is unrealistic. In this work, we follow this argument and leverage parallel data and alignments obtained from them to gain informative training signals to improve cross-lingual transfer in downstream tasks. 
For non-contextualized embeddings, alignment based techniques for word embeddings have been thoroughly surveyed in \citet{Ruder2019ASO}. %Majority of the previous works that utilize OT for alignment do so at the type level.
For contextualized embeddings, one direction of efforts to improve cross-lingual word representations is to use the Procrustes objective to project the monolingual embeddings from one language to the monolingual embedding space in another \cite{wang2019crosslingual, Schuster2019CrossLingualAO}. However, this generates a transformation matrix for each language pair which can be inconvenient to apply in downstream tasks. Another direction is to use explicit alignment objective %relying on parallel sentences learning alignments 
at the sentence level, word level, or both which allows simultaneous learning from different languages, as opposed to rotation based approaches.

Studies that depend on sentence level alignment achieve significantly high performance on bi-text sentence retrieval tasks \cite{Artetxe2019MassivelyMS, Zweigenbaum2017OverviewOT}, and by design they are not applicable to word based applications. For instance, LASER \cite{Artetxe2019MassivelyMS} learns massively multi-lingual encoder using a huge parallel corpus whereas \citet{Feng2020LanguageagnosticBS}  trains a bi-directional dual encoder with an additive softmax margin loss to perform translation ranking among in-batch examples. Similar to this line of work, we rely on only parallel sentences as external sources to fine-tune the model, but we define word alignment objective instead. 

Other studies use word alignment objective to align parallel word pairs and fine-tune the contextualized multi-lingual LM  \cite{cao2020multilingual,wu2020explicit, Nagata2020ASW}. \citet{cao2020multilingual} use regularized L2 based alignment objective to align parallel word pairs. \citet{wu2020explicit} use contrastive learning to align parallel word pairs relative to negative pairs in the batch. These approaches rely on unsupervised word aligners which are often sub-optimal to generate the parallel word pairs (e.g. FastAlign \cite{Dyer2013ASF} or optimal transport \cite{grave2018unsupervised}) and use these pairs as weak form of supervision. Our work is most similar to these methods in that we use word level alignment objective; however, we learn the aligned word pairs implicitly during optimization rather than obtaining them beforehand using external aligners and applying heuristics to keep only one-to-one mappings. 

More recently, \newcite{chi2021improving} developed an end-to-end model that first aligns both source and target words with OT and then use the alignments as self-labels to fine-tune the contextualized LM. They use three objective functions for fine-tuning:  Masked Language Modeling (MLM) \citep{devlin2019bert}, Translation Language Modeling (TLM) \cite{lample2019crosslingual}, and the cross entropy between predicted masked words and their corresponding alignments obtained from OT. Similar to their work, we use OT based signals to fine-tune the contextualized LM, but we instead use the average cost of OT alignments for fine-tuning. There are other studies that attempt to combine various objectives for learning cross-lingual supervision. For example, \citet{Dou2021WordAB, Hu2020ExplicitAO} incorporate the following objectives on cross-lingual data: MLM, TLM, sentence level alignment (e.g. parallel sentence identification objective), and word level alignment. In this paper, we do not investigate combined objective functions similar to these works. We believe that adding more objectives can further boost the performance and we leave it for future work.

%TODO cite bilingual lexicon induction paper

\section{Method}
\label{methods}
%Figure \ref{finetuning_ot} shows the overall fine-tuning process. As input, we require parallel sentences (i.e. pairs of aligned sentences in source and target languages) and contextualized multi-lingual LM. For each model iteration, we first represent source and target sentences independently with the pre-trained contextualized LM (Section \ref{word_representation}). We use that as input for OT optimization applied for each language independently (Section \ref{OT_algo}). We then fine-tune the contextualized LM with the accumulated  regularized loss values across all languages as a guidance (Section \ref{WassAlign}). We formulate the task of OT as  minimizing the cost of transferring knowledge within context, from a non-English source sentence to an English target sentence in an unsupervised fashion.\footnote{The method can be applied in any language pair (e.g. Bulgarian-Russian). We choose English since resources are available in abundance including parallel datasets and evaluation benchmarks for cross-lingual transfer.} 

Figure \ref{finetuning_ot} shows the overall fine-tuning process. As input, we require parallel sentences (i.e. pairs of aligned sentences in source and target languages) and contextualized multilingual LM. We use English as fixed target language and other non-English languages as source language (more details in Section \ref{datasets}). For each model iteration, we first embed words in source and target sentences independently with the pre-trained contextualized LM (Section \ref{word_representation}). These representations are then used as input for OT optimization applied for each source-target language pair. We then fine-tune the contextualized LM with the accumulated regularized loss across all language pairs as a guidance (Section \ref{WassAlign}). We formulate the task of OT as minimizing the cost of transferring knowledge within context, from a non-English source sentence to an English target sentence in an unsupervised fashion.\footnote{The method can be applied to any language pair (e.g. Bulgarian-Russian). We choose English since resources are available in abundance including parallel datasets and evaluation benchmarks for cross-lingual transfer.}

\subsection{Input Representation}

OT optimization is flexible to align different textual units such as words and subwords. %As illustrated in Figure \ref{word_representation_figure}, 
We provide contextualized representations for words/subwords in source and target sentences as input for the OT optimization process. We use the last layer of pretrained LMs to obtain the contextualized representations.\footnote{We also investigated other layers for word representation but empirically found the last layer to be the best.} For word representations, to have a fair comparison with \citet{cao2020multilingual}, we follow their assumption that the last subword embedding for a word contains sufficient contextual information to represent the word.\footnote{The choice of subword embedding to represent each word (e.g. first or last subword or mean pool of all subwords) can be empirical and can differ depending on the language properties. For instance, the morphological inflection in most European languages lies  in the suffix while head words of compounds tend to occur on the right in Germanic languages. Hence, the last subword representation may contain more morpho-syntactic information than head word depending on the language.} Subwords allow for more nuanced alignments and an increase in the vocabulary coverage which can be beneficial for languages that are rich with compounds or morphemes (e.g. Arabic and German).\footnote{For example, the morpheme \textit{h} in the Arabic word “ktbh” corresponds to \text{it} in the English segment “he wrote it”.}

% As illustrated in Figure \ref{word_representation_figure}, we provide contextualized representations for words in source and target sentences as input for the OT optimization process. We use the last layer of pretrained LMs to obtain contextualized word representations.\footnote{We also investigated other layers for word representation but empirically found the last layer to be the best.} To have a fair comparison, we follow the assumption made by \citet{cao2020multilingual} that the last subword embedding for a word contains sufficient contextual information to represent the word.\footnote{The choice of subword embedding to represent each word (e.g. first or last subword or mean pool ) can be empirical and can differ depending on the language properties. For instance, the morphological inflection in most European lies  in the suffix while head words of compounds tend to occur on the right in Germanic languages. Hence, the last subword representation may contain more morpho-syntactic information or head word depending on the language.} With OT, we can also benefit from its flexibility to align textual units at subword level. Subwords allow for more nuanced alignments and an increase in the vocabulary coverage which can be beneficial for languages that are rich with compounds or morphemes (e.g. Arabic and German).\footnote{For example, the morpheme \textit{h} in the Arabic word “ktbh” corresponds to \text{it} in the English segment “he wrote it”.} 

For each source-target language pair, we pass a batch of parallel sentences, represented by contextualized words/subwords embeddings as input for the OT optimization, which in turn learns to minimize the cost of transferring from source to target distributions. This process is applied to compute independent OT optimization for each source-target language pairs independently.\footnote{Combining different languages in one OT process increases the learning complexity - refer to Appendix \ref{shuffling} for more details.} We base our model on multilingual BERT (mBERT), that is jointly trained on 104 languages in which a shared vocabulary is constructed. Techniques discussed here are agnostic to the choice of pre-trained multilingual LMs.\footnote{We started with mBERT to have a fair comparison with other works that fine-tune with alignment objective.}

\label{word_representation}
% \begin{figure}
% \centering
%   \includegraphics[width=0.6\linewidth]{word_representation.png}
%   \caption{Illustration of word representation process}
%   \vspace{-5mm}
%   \label{word_representation_figure}
% \end{figure}

\subsection{Optimal Transport Optimization}
\label{OT_algo}
We use Sinkhorn divergence which interpolates between Wasserstein distance (i.e. Optimal Transport) and Maximum Mean Discrepancy (MMD), leveraging both OT geometrical properties and MMD efficiency in high-dimensional spaces \cite{ramdas2015wasserstein,feydy2019interpolating}. MMD is an energy distance or kernel which adds an entropic penalty/regularization for the optimizer and is mathematically cheaper to compute %and is less complex compared to OT 
\cite{gretton2008kernel}. We use the variant introduced in \citet{feydy2019interpolating} which leads to entropic smoothing for the weights and more stabilized and unbiased gradients as the following:
%%DONE YZ: can you elaborate on the "efficiency" advantage of OT?

\begin{small}
\vspace{-5mm}
\begin{equation}
\label{ot_align_loss}
\begin{split}
%w =  \langle (K\odot \hat{c}b, a) \rangle \\
 S_{\epsilon}(\alpha, \beta) = OT_{\epsilon}(\alpha, \beta) -  \frac{1}{2} OT_{\epsilon}(\alpha, \alpha) - \frac{1}{2} OT_{\epsilon}(\beta, \beta) \\
 OT_{\epsilon}(\alpha, \beta) = \min_{\pi} \langle \pi, C   \rangle + \epsilon KL(\pi, \alpha \otimes \beta) \\
 s.t. \; \pi>=1,  \;  \pi1=\alpha, \; \pi^T1=\beta,
\end{split}
\end{equation}
\vspace{-2mm}
\end{small}

\noindent where $\alpha$ and $\beta$ (initialized with uniform distribution) represent weights of words for each sample in the source and target  distributions, respectively.\footnote{We also found improvements in some languages with TF-IDF initialization; however, TF-IDF relies on computing statistics on the overall corpus which can be insufficient to compute such statistics for low resource languages.} Note that each ($\alpha$ and $\beta$) must sum to 1. We use Euclidean distance to encode $C$ as the ground cost in Equation \eqref{ot_align_loss}. $C$ is a $n \times m$ matrix, which represents the effort or cost of moving a point in source distribution to a point or a set of points in target distribution; $n$ and  $m$ are the number of words in source and target languages respectively. $\pi$ is also a $n \times m$ matrix denoting soft alignment between a word in source language to word(s) in target language (i.e. how much probability mass from a point in source distribution is assigned to a point in target distribution).

The $OT_{\epsilon}$ optimizer works by finding word matches between source and target sentences while minimizing the ground cost.  The $OT_{\epsilon}$ solver is controlled by  $\epsilon= 0.05$ to balance between Wasserstein distance and MMD ($KL$ term in Equation \eqref{ot_align_loss}).\footnote{We investigated few values for $\epsilon$. The default value $\epsilon= 0.05$ in Geomloss  \citep{feydy2019interpolating} provides the best results.}  To minimize this distance, we use Sinkhorn iterative matching algorithm which finds the solution of Equation \eqref{ot_align_loss} in terms of dual expression by iteratively updating the dual vectors between source and target until convergence \cite{feydy2019interpolating}. We use $\pi$ as the final alignment between words in the source and target sentences. 

Given that our approach works on contextualized embedding, where the individual word representation is different based on the context, applying OT to the entire training data is computationally prohibitive. Previous studies proposed the use of mini-batch strategy to apply OT on large scale datasets and proved its effectiveness as an implicit regularizer in machine learning settings \cite{fatras2021minibatch,fatras2020learning}. We follow the mini-batch strategy to learn OT on a batch of parallel sentences for each language pair independently  and use the resultant loss function to fine-tune our model as shown in Figure \ref{finetuning_ot}.

\subsection{Fine-tuning with OT}
\label{WassAlign}
`To fine-tune the pre-trained LM with OT, we first accumulate the cost of alignments obtained by  $S_{\epsilon}(\alpha, \beta)$ in Equation (\ref{ot_align_loss}) for each source-target language pair as discussed in Section \ref{OT_algo}. Similar to \citet{cao2020multilingual}, we additionally add a regularization term to the OT loss to  penalize the model if the target language embeddings in the tuned model shifts far from its initialization. 

%The OT alignment cost obtained for each source-target language pair (as mentioned in Section \ref{OT_algo}) is then combined and used as averaged loss value, averaged over all language pairs OT alignments. Similar to \citet{cao2020multilingual}, we additionally add a regularization term to the OT loss given by $S_{\epsilon}(\alpha,\beta)$ to  penalize the model if the target language embeddings in the tuned model shifts far from its initialization. 

\begin{small}
%\vspace{-5mm}
\begin{equation}
\label{eq_ot_regularized}
\begin{split}
l(c; P^k) = &- S_{\epsilon}^{k}(\alpha,\beta) \\
&+  \lambda \sum\limits_{t \in P^k} \sum\limits_{i=1}^{len(t)} \left\Vert c(j,t) - c_{0}(j,t) \right\Vert_2^2, 
\end{split}
\end{equation}
%\vspace{-3mm}
\end{small}

\noindent where $\lambda$ is set to $1$ and t is a target sentence in the parallel corpus $P^{k}$ for language k. $c(j,t)$ represents the contextualized representation for a word j in sentence t with the language model being tuned whereas $c_{0}(j,t)$ represents the initial representation with the un-tuned contextualized language model.  
We then back-propagate the resultant regularized loss (as shown in Equation \eqref{eq_ot_regularized}), summed over all $K$ languages, i.e., $L(c) = \sum\limits_{i=1}^{K} l(c; P^{i})$ to fine-tune the contextualized word representations.

%In particular, since we accumulate the OT losses from each language pair before updating contextualized LM, different languages used during training can influence each other's representations.   

\section{Experimental Setup}
\label{setup}
\subsection{Data Pre-processing}
\label{datasets}
Following previous studies \cite{lample2019crosslingual, cao2020multilingual}, we use parallel data (approximately 32M sentence pairs) from a variety of corpora to cover different language pairs and domains as shown in Appendix \ref{parallel_corpus_details} - %Table \ref{table:data_stats} shows the data sources and number of parallel sentences for language pairs considered in our experiments. We use a variety of corpora to cover different language pairs and domains: 
 Europarl corpora \cite{koehn2005europarl}, MultiUN \cite{eisele-chen-2010-multiun}, IIT Bombay \cite{kunchukuttan2018iit}, Tanzil and GlobalVoices \cite{tiedemann-2012-parallel}, and OpenSubtitles \cite{lison-tiedemann-2016-opensubtitles2016}. In all cases, we use English (en) as the target language and the tokenizer in \citet{koehn-etal-2007-moses}. We use 250K sentences for training, upsampling from language pairs where this much data is not available. We shuffled the data to break their chronological order if any. For our main model, we consider the following five languages:  
Bulgarian (bg), German (de), Greek (el), Spanish (es), and French (fr), similar to \citet{cao2020multilingual}. For our larger model, we additionally used the following languages: Russian (ru), Arabic (ar), Mandarin (zh), Hindi (hi), Thai (th), Turkish (tr), Urdu (ur), Swahili (sw), and Vietnamese (vi). 

\subsection{Model Optimization}
\label{model_optimization}
We use Adam \citep{kingma2017adam} for fine-tuning pre-trained LM using OT with learning rate of $5e-5$ for one epoch. We sample equal-sized parallel sentences from each language pair, do a forward pass accumulating losses for each language pair and then backpropagate based on combined loss from all language pairs. We use Geomloss for Sinkhorn divergence with its default parameter values \citep{feydy2019interpolating}. %\footnote{We observe that learning saturates if we train for more than one epoch with 250K parallel sentences.}
We empirically chose batch size of 24 and gradient accumulation step of 2 to balance between speed, memory, and model accuracy.\footnote{Roughly, batch size $=1$ takes at least 5 days to complete fine-tuning while batch size $=24$ takes around 8 hours on a single NVIDIA V100 GPU.} Having smaller batch sizes or updating the gradients too frequently slightly hurt the performance and may lead to over-fitting the contextualized LM to noisy parallel sentences or irregular patterns. %We reach memory capacity with higher batch sizes.

\subsection{Evaluation}
We evaluate our proposed method for two tasks provided by XTREME benchmarks \cite{hu2020xtreme}: XNLI for textual entailment where the task is to classify the entailment relationship between a given pair of sentences into entailment/neutral/contradiction  \cite{conneau2018xnli,williams2018broadcoverage}; XQuAD for question answering where the task is to identify the answer to a question as a span in the corresponding paragraph \cite{artetxe2020crosslingual,rajpurkar2016squad}.\footnote{Refer to \cite{hu2020xtreme} for more details regarding these benchmarks. We use XTREME open source code implementation -  \url{https://github.com/google-research/xtreme}} These tasks evaluate zero shot transferability and hence we train all tasks using English labelled data with cross-entropy loss and test on the target languages. More details about the task settings can be found in Appendix \ref{hyperparameter_settings}. To measure the improvements, we use F1 score for textual entailment; F1 and EM (Exact Match) scores for question answering which reflect the partial and exact matches between the prediction and ground truth, respectively. 

\subsection{Models Comparison}
In addition to mBERT, we compare our approach to the following baselines:  \begin{enumerate*}
    \item XLM \cite{lample2019crosslingual} which use similar objective as mBERT with a larger model and vocabulary,
    \item L2 \cite{cao2020multilingual}  which uses L2 based alignment objective,
    \item AMBER \cite{hu2021explicit} for XNLI which uses a combination of MLM, TLM, word alignment and sentence alignment objectives,\footnote{We compare our model with the published AMBER variant that does not use sentence alignment as that is most comparable to our settings.} 
    \item MAD-X \cite{pfeiffer2020madx} for XQuAD which leverages language and task adapters for efficient cross lingual transfer.\footnote{We internally reproduce MAD-X scores with mBERT as the main model to show fair comparison with our method. MAD-X\textsuperscript{base} and MAD-X\textsuperscript{mBERT} refers to MAD-X architectures with XLMR-Base and mBERT as main model respectively.}
\end{enumerate*}
We also compare how our model performs with respect to current state-of-the-art model i.e. XLMR \cite{Conneau2020UnsupervisedCR} which is same as XLM but trained on much more data.

\begin{table}[h!]
\small
\centering
\addtolength{\tabcolsep}{-2.0 pt}
%\parbox{.45\textwidth}{
\begin{tabular}{l | c c c } 
\hline
& \textbf{XNLI} & \multicolumn{2}{c}{\textbf{XQuAD}} \\
\textbf{Model} & \textbf{F1} & \textbf{F1} & \textbf{EM} \\
\hline
 mBERT & 71.9 & 73.1 & 57.0 \\
 XLM & 74.6 & 66.5 & 50.2\\
 AMBER & \textbf{76.4} & - & - \\
\hline
mBERT$\dagger$ & 73.5 & 73.4 & 57.8 \\
L2$\dagger$ & 74.6 & 68.0 & 51.6\\
MAD-X\textsuperscript{mBERT}$\dagger$ & - & 70.2 & 53.8 \\
\hline
%WordOT & \underline{75.4} & \underline{\textbf{74.7}} & \underline{\textbf{59.0}} \\
WordOT (Ours) & 75.4 & \textbf{74.7} & \textbf{59.0} \\

\hline
\end{tabular}%} 

\caption{Averaged scores for XNLI and  XQuAD benchmarks across three runs compared to baselines in seen languages. \textbf{Bold} scores are the highest in the respective columns. $\dagger$ refers to internal benchmarking, where we either obtained the models from the authors or implemented internally.}
\vspace{-5mm}
\label{table:seen_xnli_xquad_results}
\end{table}

\section{Results and Discussion}
\label{results}
Table \ref{table:seen_xnli_xquad_results} shows the performance of our proposed method (WordOT) averaged for languages that are seen during OT fine-tuning. We compare that to the baselines and state-of-the-art approaches in the respective evaluation tasks (XNLI and XQuAD).  We run all tasks for three seeds for each considered language and report the average scores for experiments that we run internally. More detailed results per language can be seen in Appendix \ref{detailed_per_language}.

\begin{table*}[h!]
\setlength{\tabcolsep}{4pt} %% default is 6pt
\small
\centering
\resizebox{\textwidth}{!}{%
\begin{tabular}{l | c c c c c c | c c c c c c c c c | c} 
\hline
\multicolumn{17}{c}{\textbf{XNLI}} \\
\hline
\multicolumn{1}{c|}{Model} & en & bg & de & el & es & fr & ar & hi & ru & sw & th & tr & ur & vi & zh & Average \\
\hline 
mBERT&	\textbf{82.6}  &	69.3&	72.0&	67.7&	75.2&	74.4 &	66.0 &	60.8&	69.4&	51.0 &	55.3&	62.9&	58.5&	71.0 &	69.9 & 73.5/67.1 \\
L2	&81.0&		73.3&	72.8&	\textbf{70.9} &	74.9&	74.4& 62.4&	59.2&	67.3&	42.8&	48.5&	54.6&	56.3&	69.7&	69.5 & 74.6/65.2 \\
WordOT&	82.1&		\textbf{73.6}&	\textbf{73.7}&	70.6&	\textbf{76.7} &	\textbf{75.9} &	66.3& 61.3&	69.7&	49.8&	54.8&	61.7&	59.4&	70.9&	70.5 & \textbf{75.4}/67.8\\
WordOT*&	81.5&		73.5&	73.3&	70.8&	76.3&	75.3& 66.0 &	61.9 &	69.7&	48.3&	55.7&	61.3&	59.7&	71.1&	\textbf{71.2} &75.1/67.7 \\
LargeWordOT&	81.8&		72.6&	73.1&	69.8&	76.1&	75.3& \textbf{66.9} &	\textbf{64.8}&	\textbf{70.4} &	\textbf{62.2} &	\textbf{60.1}&	\textbf{68.6}&	\textbf{60.6}&	\textbf{72.2} &	70.4 & 74.8/\textbf{69.7}\\
\hline
\end{tabular}
}
\quad
\setlength{\tabcolsep}{4pt} %% default is 6pt
\begin{tabular}{l | c | c c c c | c c c c c c c | c} 
\multicolumn{14}{c}{\textbf{XQuAD}} \\
\hline
\multicolumn{1}{c|}{Model} & Metric &en&	de&	el	&	es & ar	&	hi&		ru	&	th	&	tr	&	vi	&	zh  & Average \\
 \hline
mBERT& \multirow{5}{*}{F1} &	83.7&		72.0 &	62.3&	\textbf{75.6} & \textbf{61.5}&	57.3&	\textbf{70.7}&	\textbf{42.2}&	54.1&	68.6&	\textbf{59.3}  &73.4/\textbf{64.3}\\
L2	&&81.4 &	67.5&	56.6&	66.2 &	48.0 & 	42.6&	62.6&	25.3&	39.0&	59.8&	48.4 &68.0/54.3\\
WordOT&&	\textbf{84.2}&		\textbf{73.6}&	\textbf{65.6} &	75.5& 58.6&	55.7&	68.6&	42.1&	51.8&	69.0&	57.3 & \textbf{74.7}/63.8\\
WordOT*&&	83.5&		72.7&	64.7&	74.2& 58.3&	53.6&	68.4&	42.0&	50.7&	68.2&	56.8 &73.8/63.0\\
LargeWordOT&&	83.4&		71.8&	63.2&	73.8& 55.9&	\textbf{59.5}&	68.9&	38.9&	\textbf{59.2}&	\textbf{70.2}&	51.4&73.1/63.3\\
\hline

mBERT & \multirow{5}{*}{EM} &	\textbf{72.4}&		55.9&	45.3&	\textbf{57.5}  & \textbf{45.5}&	44.1&	\textbf{53.9} &	32.6&	39.5 &	\textbf{49.7} &	\textbf{49.7} &57.8/\textbf{49.6} \\
L2 & 	&69.4&		51.3&	40.3&	45.4& 29.9&	28.4&	44.1&	17.8&	25.5&	41.2&	40.0 &51.6/39.4\\
WordOT& &	\textbf{72.4}&		\textbf{57.8}&	\textbf{48.5}&	57.1& 40.9&	40.8&	51.3&	32.9&	37.1&	49.1&	48.6 & \textbf{59.0}/48.8\\
WordOT* & 	&71.7&		56.9&	47.6&	55.5& 40.4&	38.8&	50.5&	\textbf{33.4} &	36.1&	48.5&	48.2 &57.9/48.0\\
LargeWordOT& &	71.8&		56.4&	46.5&	55.6	& 37.1& \textbf{44.9}	&50.8	&30.1	&\textbf{44.4} &	49.3&	43.1&57.6/48.2\\

\hline
\end{tabular}
\caption{F1 score in XNLI and (F1 / EM) scores in XQuAD for each language  across three runs.  \textbf{Bold} scores are the highest in the respective column and metric. \textit{Average} starts with the average score for the 5 seen languages (separated by vertical bar) in L2, \textit{WordOT} and \textit{WordOT*} followed by the average score for the 15 languages seen during optimization in \textit{LargeWordOT}, separated by /. }
\vspace{-5mm}
\label{table:large_results}
\end{table*}

Compared to the baseline mBERT, we obtain +1.9\% and +1.3\% F1 scores on average in XNLI and XQuAD, respectively. Compared to L2 \cite{cao2020multilingual}, we obtain an average improvement of +0.8\% for XNLI and +6.7\% for XQuAD in F1 scores. In XNLI, we obtain comparable F1 score  (-1.0\%) to the more recent model - AMBER \cite{hu2021explicit}. This could be attributed to the TLM \cite{lample2019crosslingual} objective used in AMBER which provides additional cross-lingual signal and hence, further boosts the performance. In XQuAD, we obtain better F1 score (+4.5\%) than the more recent work - MAD-X \cite{pfeiffer2020madx} - showing the effectiveness of our method.

\paragraph{More languages during optimization:} In our previous results, we fine-tuned mBERT with parallel sentences drawn from a set of 5 languages (refer to Section \ref{datasets}). We also investigate whether adding more languages during fine-tuning with OT (LargeWordOT) would help improve the performance. %To do so, 
We expanded the set of languages to all 15 languages as described in Section \ref{datasets} (also Table \ref{table:data_stats} in Appendix \ref{parallel_corpus_details}). As a result of computational complexity of OT, we instead used batch size of 8 and gradient accumulation step of 3 to overcome memory overhead. We also re-trained the model with previous 5 languages using new hyper-parameter settings (WordOT*) to have a fair comparison between both models. 

Table \ref{table:large_results} shows the results for XNLI and XQuAD, respectively. In XNLI, we obtain 2.6\% improvements with LargeWordOT compared to mBERT. We do not observe improvements on average for XQuAD benchmark for LargeWordOT. This could be a byproduct of fine-tuning mBERT with parallel texts of different languages, exposing their similarities as well as their differences to the whole network. XQuAD, being a difficult task compared to XNLI is impacted more by these differences in languages' properties (language family, writing script, word order etc.). Moreover, we observe that adding more languages during fine-tuning slightly decreases the average score for the 5 languages seen as in WordOT*. Looking at scores for each language individually, we gain significant improvements for hi, sw, and tr across the two tasks. Note that the monolingual data available in Wikipedia is scarce for sw, hi, tr, and ar.  

We also examine the impact of OT fine-tuning on unseen languages from the performance of WordOT*. We notice similar or better performance compared to LargeWordOT on average for all languages for both tasks, thereby showing that the performance on remaining languages on average is comparable. In addition, Table \ref{table:large_results} shows that WordOT performs overall better than its counterpart (WordOT*) both of which differ in the batch size (24 vs. 8) and the number of gradient accumulation steps (2 vs. 3). Hence, we presumably would obtain better scores with higher batch size for OT if the implementation is optimized for memory efficiency.

\paragraph{Notes on OT Efficiency:} To examine the  efficiency of our proposed method, we computed the time taken by one epoch of fine-tuning mBERT with five language pairs (250K parallel sentences for each pair). On a single NVIDIA V100 GPU, it took approximately $8$ hours to complete one epoch, which is relatively 30\% higher compared to L2 based alignment method which took approximately $6$ hours with the same settings. This increase is expected as our method considers every combination of words from source to target in order to find OT mapping with minimum cost for each step of fine-tuning. Hence, it performs at least $O(n*m)$ operations, where $n$ and $m$ are the number of  words in source and target languages, respectively. On the other hand, L2 based alignment considers only precomputed one-to-one mapping which speeds up the process. This is a trade-off between time and accuracy where OT outperforms L2 in both tasks for seen and unseen languages in terms of accuracy. The time complexity only impacts the model during fine-tuning which is done once.

\vspace{-2.5mm}
\paragraph{Word vs. subword alignments:} As discussed in Section \ref{word_representation}, OT is flexible to align different textual units. We compare between fine-tuning at word level (WordOT) and subword level (SubOT). Table \ref{table:subword_word_results} shows that SubOT slightly improves the scores in XNLI and slightly decreases the scores in XQuAD. We observe  individual improvements for some languages with subword level alignment. In XNLI, the F1 scores for Greek and German slightly increased by 0.65\% and 0.47\% respectively with subword information. Both languages exhibit compounding structure as opposed to remaining languages seen during training in which the benefit is less observed (\textless  0.29\%). For XQuAD, we observe slight drop in overall performance with subword information.\footnote{We observe benefits for some low resource languages such as th which improved +2.4\% F1 and +1.6\% EM.} This can be attributed to the nature of XQuAD task in which a span of information is identified. We believe that the difference between word and subword can be more pronounced when we construct language specific vocabulary and/or increase the vocabulary capacity. 

\begin{table}[!h]
\small
\centering
\begin{tabular}{c | c c c c} 
\hline
\multirow{2}{*}{\textbf{Model}} & \multicolumn{2}{c}{\textbf{XNLI}}  & \multicolumn{2}{c}{\textbf{XQuAD}}\\
& seen & all & seen & all \\
\hline
WordOT & 75.4 & 67.8 &  \textbf{74.7}/\textbf{59.0}  & \textbf{63.8}/\textbf{48.8} \\
SubOT  & \textbf{75.7} &	\textbf{67.9} & 74.0/58.3  &63.5/48.5 \\
%WordOT & 74.1 & 67.9 &  \textbf{71.4} / \textbf{54.5} & 63.8 / 48.8  \\
%SubOT  & \textbf{74.7} &	\textbf{68.2} & 71.2 / 54.2 & \textbf{63.9} /	\textbf{48.9} \\
\hline
\end{tabular}
\caption{Scores (F1 for XNLI F1 / EM for XQuAD) for SubOT vs. WordOT. \textit{``All''} represents the average of both seen and unseen languages during optimization.}
\vspace{-3mm}
\label{table:subword_word_results}
\end{table}

\vspace{-0.5mm}
\paragraph{Impact of amount of parallel data:} In all previous experiments, we used 250k parallel sentences (upsampled if needed). %for language pairs not having this much data)
Adding more language pairs during training with OT increases the fine-tuning time thus limiting the scalability of our proposed approach. In addition, the impact of OT if we have limited amount of parallel data for a low resource language is not clear.\footnote{Low resource languages here refer to languages covered by mBERT vocabulary but has limited amount of parallel data when using our approach.}
To address the aforementioned two points, we investigate the impact of reducing the amount of available parallel data. These experiments were performed using LargeWordOT. We can see from Table \ref{table:diff_amount_parallel_results_xnli} that for XNLI, we can achieve comparable performance (-0.4\% absolute) with as low as 50k sentences, i.e. one-fifth of the data. Similar experiments for XQuAD can be found in the Appendix \ref{parallel_data_xquad}.
%As shown in Table \ref{table:diff_amount_parallel_results}, XNLI results show that we can achieve comparable performance (-0.5\%) with as low as 10k parallel sentences.
%\footnote{Similar experiments for XQuAD can be found in the appendix.}.
This shows that alignment using OT is robust to low data scenarios, especially for languages where huge amounts of parallel data might not be available.

%SAWSAN: are these scores for development or test? how many epochs?

\begin{table}[h!]
\small
\centering
\addtolength{\tabcolsep}{-2.0 pt}
\begin{tabular}{l | c c c c c c | c } 
\hline
\multicolumn{8}{c}{\textbf{XNLI}} \\
\hline
Model & en & bg & de & el & es & fr & Avg \\
\hline
 mBERT & \textbf{82.6} & 69.3 & 72.0 & 67.7 & 75.2 & 74.4 & 73.5 \\
 1k & 82.3 & 69.9 & 72.5 & 67.3 & 75.0 & 74.6 & 72.7 \\
10k & 81.7 & 71.9 &	72.2 & 68.5 &	75.5&	74.6 & 74.1\\
50k & 81.4 & \textbf{72.7} &	72.7 & 69.2 &	75.8&	74.7 & 74.4\\
250k &  81.8 & 72.6 & \textbf{73.1} & \textbf{69.8} &	\textbf{76.1} & \textbf{75.3} & \textbf{74.8}\\
\hline
\end{tabular}

% \quad
% \addtolength{\tabcolsep}{-2.5 pt}
% \begin{tabular}{l | c c c c | c } 
% \multicolumn{6}{c}{\textbf{XQuAD}} \\
% \hline
% Model & en & de & el	&	es	& Avg \\  
% \hline 
% mBERT & 82.4/70.6 & 71.1/54.8 & 60.5/44.9 & 74.8/56.5 & 72.2/56.7 \\
% %& & & & &\\
% 10k &	83.8/72.9 & 70.8/55.9  & 62.9/45.9 &	73.8/55.5  & 72.8/57.5 \\
% %& & & & &\\
% 50k &	83.3/71.1 & 72.0/57.3 &	62.7/44.9 &	75.6/56.9 &	 73.4/57.5  \\
% 250k &	83.2/71.3 & 71.1/56.1 &	62.9/46.2 &	74.3/56.7 &	 72.9/57.6  \\
% \hline
% \end{tabular}
\caption{XNLI F1 scores  %and XQuAD scores (F1 / EM) 
for different amounts of parallel data. mBERT represents the case where we have no parallel datasets}
\label{table:diff_amount_parallel_results_xnli}
\vspace{-5mm}
\end{table}

\paragraph{State-of-the-art Comparison:}
We compare our method to the state-of-the-art model XLMR which has a larger capacity in terms of model and/or training data sizes. Due to efficiency reasons, we apply  OT on XLMR\textsuperscript{base} which has similar model size compared to mBERT but is trained on significantly larger amount of data (2.5TB) and larger vocabulary.\footnote{OT can also be applied in XLMR\textsuperscript{large}; however, this would require parameter tuning to overcome memory issues.} As shown in Table \ref{table:seen_xnli_xquad_results_sota}, we observe comparable or slightly lower results when we apply OT on XLMR\textsuperscript{base}. Hence, explicit alignment objective with OT as our proposed method did not help further boost the performance; this is in line with the findings of \citet{wu2020explicit} which show improvements for different alignment objectives over mBERT but not XLMR. 

We speculate that the robustness of XLMR over alignment objectives can be attributed to the large amount of data used for pre-training even for low resource languages. Hence, to further boost the performance, there must be consideration for the amount of data used for alignment in correlation with the pretrained data (e.g. mBERT shows benefits from our method with even smaller size of data, i.e. 50K samples). In addition, the definition of alignment objective is a determining factor. For example, \citet{chi2021improving} showed improvements when they used OT based alignment as self-labels to minimize the loss between predicted masked word and the corresponding aligned word. Note that \citet{chi2021improving} also uses large amount of data for training.

\begin{table}[h!]
\small
\centering
\addtolength{\tabcolsep}{-2.0 pt}
\begin{tabular}{l | c c c } 
\hline
 & \textbf{XNLI} & \multicolumn{2}{c}{\textbf{XQuAD}} \\

\textbf{Model} & \textbf{F1} & \textbf{F1} & \textbf{EM} \\
\hline
%XLMR & \textbf{81.6} & \textbf{82.2} & \textbf{66.2} \\
%XLMR\textsuperscript{base} & 78.4 & 76.7 & 60.8 \\
%WordOT & 75.4 & 74.7 & 59.0 \\
XLMR & \textbf{84.1} & \textbf{82.2} & \textbf{66.0} \\
XLMR\textsuperscript{base} & 77.5 & 77.0 & 61.4 \\
WordOT\textsuperscript{base} & 77.6 & 76.4 &	60.8\\
\hline
\end{tabular}%} 
\caption{Comparison with state-of-the-art (XLMR). In WordOT\textsuperscript{base}, we apply OT based fine-tuning on XLMR\textsuperscript{base}. \textbf{Bold} scores are the highest in the respective column. All results were obtained internally and are averaged across three runs. For learning rate, we use $5e-6$ for XLMR evaluation benchmarks.} 
\vspace{-5mm}
\label{table:seen_xnli_xquad_results_sota}
\end{table}

\section{Qualitative Analysis for OT}
\label{analysis}
Our objective is not to obtain explicit word alignment but rather compute the cost of transferring both distributions to each other and use this cost to guide the fine-tuning process. We examine the obtained alignments during fine-tuning for two language pairs (German-English and Arabic-English) to inspect potential errors. We found that alignments are capable to include word relationships other than one-to-one mapping. For instance, the German compound nouns \textit{``Vorsichtsprinzip''} and \textit{``Rahmengesetzgebung''} are correctly aligned to \textit{``precautionary approach''} and \textit{``framework legislation''}.
%For instance, the Arabic word \<الاتفاقية> is aligned to \textit{“the convention"} in English. The German words \textit{“Vorsichtsprinzip"} and \textit{“Rahmengesetzgebung"} are aligned to \textit{“precautionary approach"} and \textit{“framework legislation,"} respectively. 
In addition, alignments do not necessarily include semantics but also highlight similar or dependent words in context, thus capturing contextual alignments.
%sometimes form a sort of clusters of similar and/or dependent words. 
For instance, in the Arabic phrase \<التدخل العسكري>, the first word \<التدخل> is aligned with its literal translation \textit{``intervention''} while \<العسكري>
is aligned with the phrase \textit{``armed intervention''}, where \textit{``arms''} is the literal translation while \textit{``intervention''} is the dependent word.
More examples in Tables \ref{examples_arabic} and \ref{german_examples} in Appendix \ref{example_section}.
%Another example is the Arabic phrase \<بيانه\ الختامي>
%where the first word \<بيانه> 
%is mapped with the meaning of its morphological suffix \textit{“his"} while 
%\<الختامي> is aligned with
% the whole phrase meaning \textit{“closing statements"}. This is a potential sub-sequence of using the last subword to represent the whole word. 
% @Sawsan: Last line is not clear "This is a potential sub-sequence of using the last subword to represent the whole word."

%In addition, each word in the Arabic phrase \<اسناد الاسبقية>  is mapped to the same words \textit{compatibility} and \textit{precedence} although ; the latter Arabic word is also mapped to \textit{the} (definite article).

%GL: What do we mean by errors here? XNLI/XQuad errors? Errors for OT alignments? Should we quantify this claim ?
%we need annotation to quantify - Yi annotated some examples - I guess it is 14 examples 
\vspace{-0.1mm}
OT as an unsupervised aligner generates incorrect alignments for some cases which could be related to quality of parallel sentences or limitations of the OT variant that we used. Some parallel sentences are not translations of each other (refer to Table \ref{table:bad_alignment_examples} in Appendix \ref{example_section}) which has a negative impact on OT especially given that we use uniform distribution which leads to finding at least one target word for each word in the source sentence. For the OT limitations, the alignments happen at the point level regardless of the word order or syntactic structure of the sentence. This indicates that a word in the source language may be aligned with more than one occurrence of the same word. For instance, the Arabic word  \textit{\<ساعة>} is mapped to the two occurrences of \textit{``hours''} in the target neglecting the clause structure. %The number of hours of subsidized childcare for under-fives had been raised from 37 hours to 50 hours per week . %وزيد عدد ساعات رعاية الطفل دون الخامسة من عمره والتي تصرف عنها إعانة , من 37 ساعة في الأسبوع إلى 50 ساعة .
This also led the model to align different morphological variants to the same instance. For example, the Arabic word \textit{\<تيسير>} is  aligned with both \textit{``facilitate''} and \textit{``Facilitating''} in the corresponding English sentence.

\section{Conclusion}
\label{conclusion}
In this paper, we investigated OT to align the space of contextualized embeddings of a source and a target sentence in order to improve contextualized word embeddings for cross-lingual settings. We trained an independent OT per language pair and used the resultant cost as a guidance to fine-tune the contextualized LM and encourage the alignment of the corresponding contextual embeddings.
We obtain improvements in sentence level evaluation tasks: XNLI and XQuAD.  As an improvement for our proposed technique, we intend to use different variants of OT such as Goromov-Wasserstein which performs the same logic presented in this paper in addition to its ability to align embeddings of different spaces, mapping both geometry and points of different embedding spaces. We would also like to combine more cross-lingual objectives using additional signals and perform evaluation on more tasks and languages. 

\section*{Acknowledgements}
We would like to thank the Lex Science team at Amazon Web Services AI for the helpful discussions; and the reviewers for providing insightful feedback. We would also like to express gratitude to Steven Cao for sharing his implementation and model in order to help us build upon his work; and clarifying our queries.  

\bibliography{anthology,custom}
\bibliographystyle{acl_natbib}

\newpage
\appendix
\section{Parallel Corpus Details}
\label{parallel_corpus_details}
\begin{table}[h!]
\small
\centering
\setlength\tabcolsep{2pt}
\begin{tabular}{l c c |l c c} 
    \multicolumn{1}{c}{Data Source} & Lang &	\#Pairs  & \multicolumn{1}{c}{Data Source} & Lang & \#Pairs \\
	\hline 
	\multirow{5}{*}{\parbox{1.5 cm}{Europarl corpora}} & bg-en &	371K & IIT Bombay & hi-en &	618K \\\cline{4-6}
	&de-en&	1M &  \multirow{3}{*}{OpenSubtitles} & th-en &	278K\\
	&el-en & 157K & &  tr-en&	9M \\
	&es-en&	990K & & 
	vi-en &	608K\\ \cline{4-6}
	&fr-en&	1M & Tanzil & 
    ur-en&	422K \\ \cline{4-6}
	\cline{1-3}
	\multirow{3}{*}{MultiUN} &ar-en& 2M & \multirow{3}{*}{\parbox{1.6cm}{Tanzil  Global-Voices}} & sw-en & 128K\\
	& ru-en & 9M & & \\
	& zh-en& 3M & & \\
	\hline
% 	\multirow{4}{*}{Tanzil} &
% 	th-en&	278K\\
% 	& tr-en&	9M\\
% 	& sw-en&	128K\\
% 	& ur-en&	422K\\
% 	\hline
% 	open subtitles & vi-en&	608K\\
% 	\hline
% Total	& & 	~32M\\
% \hline
\end{tabular}
\caption{The data source and number of parallel sentences in each pair of languages. Overall ~32M parallel sentences combined}
\vspace{-3mm}
\label{table:data_stats}
\end{table}

\section{Task Hyperparameter Settings}
\label{hyperparameter_settings}
We benchmarked the performance of our model and baselines with XNLI and XQuAD datasets using the same settings as XTREME \cite{hu2020xtreme}. However, for internally implemented MAD-X using XLMR-base or mBERT as the base model, we followed the XQuAD scripts as in \cite{pfeiffer2020AdapterHub} because of incompatibility in versions of certain packages between XTREME\footnote{\url{https://github.com/google-research/xtreme}} and Adapters\footnote{\url{https://github.com/Adapter-Hub/adapter-transformers}} libraries. We used learning rate of 1e-4 for adapters and trained on XQuAD task for 4 epochs with a batch size of 4 and
gradient accumulation steps of 4. Rest of the settings were similar to as mentioned in \citet{pfeiffer2020madx}, i.e., adapter sizes correspond to reductions of 2 for language adapters, 2 for invertible adapters, and 16 for task adapters.   

%SAWSAN: is the above setting for QA different to what we are using for remaining models? 

\section{Detailed Results Per Language}
\label{detailed_per_language}
Table \ref{table:seen_xnli_xquad_detailed_results} shows comparison of our method with baselines and state-of-the-art approaches per language (average numbers across 3 runs). 

\begin{table*}[h!]
\small
\centering
% \addtolength{\tabcolsep}{-2.0 pt}
%\parbox{.45\textwidth}{
\begin{tabular}{l c c c c c c | c } 
\hline
\multicolumn{8}{c}{\textbf{XNLI}} \\
\hline
\textbf{Model} & \textbf{en} & \textbf{bg} & \textbf{de} & \textbf{el} & \textbf{es} & \textbf{fr} & \textbf{Avg} \\
\hline
 mBERT & 80.8 &	68.0 & 70.0 & 65.3 & 73.5 & 73.4 & 71.9 \\
 XLM & 82.8 & 71.9 & 72.7 & 70.4 & 75.5 & 74.3 & 74.6 \\
 XLMR & \textbf{88.7} & \textbf{83.0} & \textbf{82.5} & \textbf{80.8} & \textbf{83.7} & \textbf{82.2} & \textbf{81.6} \\
 %\multirow{2}{*}{\parbox{2 cm}{AMBER}} & 84.1 & 73.9 & 74.7 & 71.6 & 76.6 & 77.7 & 76.4 \\
% & & & & & & & \\
%& & & & & & & \\
 AMBER & 84.1 & 73.9 & 74.7 & 71.6 & 76.6 & 77.7 & 76.4 \\
\hline
%mBERT$\dagger$ & \underline{82.6} & 69.3 & 72.0 & 67.7 & 75.2 & 74.4 & 73.5\\
mBERT$\dagger$ & 82.6 & 69.3 & 72.0 & 67.7 & 75.2 & 74.4 & 73.5\\

%L2$\dagger$ & 81.0 & 73.3 &	72.8 & \underline{70.9} &	74.9 &	74.3 & 74.6 \\
L2$\dagger$ & 81.0 & 73.3 &	72.8 & 70.9 &	74.9 &	74.3 & 74.6 \\

\hline
\multicolumn{8}{c}{\textbf{Ours}} \\
\hline
%WordOT & 82.1 & \underline{73.6} &	\underline{73.7} &	70.6 & \underline{76.7} &	\underline{75.9} & \underline{75.4} \\
WordOT & 82.1 & 73.6 &	73.7 &	70.6 & 76.7 &	75.9 & 75.4 \\

\hline
\end{tabular}%} 
\quad
\addtolength{\tabcolsep}{-2.0 pt}
%\parbox{.2\textwidth}{
\begin{tabular}{l c c c c | c } 
%\smaller[3]
\\
\hline
\multicolumn{6}{c}{\textbf{XQuAD}} \\
\hline
\textbf{Model} & \textbf{en} & \textbf{de} &	\textbf{el}	&	\textbf{es}	& \textbf{Avg} \\  
\hline
mBERT & 83.5/72.2 & 70.6/54.0 & 62.6/44.9 & 75.5/56.9 & 73.1/57.0 \\
XLM & 74.2/62.1 & 66.0/49.7 & 57.5/39.1 & 68.2/49.8 & 66.5/50.2 \\
XLMR & \textbf{86.5}/\textbf{75.7} & \textbf{80.4}/\textbf{63.4} & \textbf{79.8}/\textbf{61.7} & \textbf{82.0}/\textbf{63.9} & \textbf{82.2}/\textbf{66.2} \\
MAD-X\textsuperscript{base} & 83.5/72.6 & 72.9/56 & 72.9/54.6 & 75.9/56.9 & 76.3/60.0 \\

\hline 
%mBERT$\dagger$ & 83.7/\underline{72.4} & 72.0/55.9 & 62.3/45.3 & \underline{75.6}/\underline{57.5} & 73.4/57.8 \\
mBERT$\dagger$ & 83.7/72.4 & 72.0/55.9 & 62.3/45.3 & 75.6/57.5 & 73.4/57.8 \\
L2$\dagger$  & 81.4/69.4 & 67.5/51.3 & 	56.6/40.3 & 66.2/45.4 & 68.0/51.6 \\
MAD-X\textsuperscript{base}$\dagger$ &  82.0/71.1 & 72.1/54.5 & 71.7/53.7 & 74.3/55.7 & 75.0/58.8 \\

MAD-X\textsuperscript{mBERT}$\dagger$ & 81.7/69.7 & 68.6/52.1 & 58.6/41.3 & 71.8/51.9 & 70.2/53.8 \\

\hline
\multicolumn{6}{c}{\textbf{Ours}} \\
\hline
%WordOT & \underline{84.2}/\underline{72.4} & \underline{73.6}/\underline{57.8} &	\underline{65.6}/\underline{48.5} &	75.5/57.1 & \underline{74.7}/\underline{59.0} \\
WordOT & 84.2/72.4 & 73.6/57.8 &	65.6/48.5 &	75.5/57.1 & 74.7/59.0 \\

\hline

\end{tabular}
%}
\caption{Averaged F1 and F1/EM scores for XNLI and  XQuAD benchmarks across three runs in seen languages. \textbf{Bold} scores are the highest in the respective columns. $\dagger$ refers to internal benchmarking, where we either obtained the models from the authors or implemented internally.} %\underline{Underlined} scores are the highest among our models internal scores (marked with $\dagger$).} 
\label{table:seen_xnli_xquad_detailed_results}
\end{table*}

\section{Impact of Amount of Parallel Data for XQuAD}
\label{parallel_data_xquad}
Table \ref{table:diff_amount_parallel_results_xquad} shows the impact of amount of parallel sentence pairs used during fine-tuning with OT for XQuAD benchmark. From the XQuAD results, we don't see a clear trend of decreasing performance with the decrease in parallel data used for OT fine-tuning. Results are more or less comparable to the baseline, with surprisingly best performance being seen with only 1k parallel sentence pairs. This could be attributed to the fact that these experiments were run using LargeWordOT that utilized 15 languages and with additional parallel data or more fine-tuning, XQuAD is being impacted negatively by the differences in these 15 languages. 

\begin{table}[h!]
\small
\centering
\addtolength{\tabcolsep}{-4.0 pt}
\begin{tabular}{l |c|c|c |c| c } 
\hline
\multicolumn{5}{c}{\textbf{XQuAD}} \\
\hline
Model & en & de & el & es & Avg \\
\hline
 mBERT & 83.7/72.4 & 72.0/55.9 & 62.3/45.3 & 75.6/57.5 & 73.4/57.8 \\
 1k & 84.4/73.3 & 72.7/56.4 & 63.9/46.8 & 75.2/56.6 & 74.1/58.3 \\
10k & 84.1/73.1 & 72.2/56.6 & 61.9/44.6 & 75.4/57.0 & 73.4/57.8\\
50k & 84.0/72.2 & 72.2/57.1 & 63.3/45.4 & 74.7/56.3 & 73.6/57.8\\
250k & 83.4/71.8 & 71.9/56.5 & 63.2/46.5 & 73.8/55.6 & 73.1/57.6\\
\hline
\end{tabular}

% \quad
% \addtolength{\tabcolsep}{-2.5 pt}
% \begin{tabular}{l | c c c c | c } 
% \multicolumn{6}{c}{\textbf{XQuAD}} \\
% \hline
% Model & en & de & el	&	es	& Avg \\  
% \hline 
% mBERT & 82.4/70.6 & 71.1/54.8 & 60.5/44.9 & 74.8/56.5 & 72.2/56.7 \\
% %& & & & &\\
% 10k &	83.8/72.9 & 70.8/55.9  & 62.9/45.9 &	73.8/55.5  & 72.8/57.5 \\
% %& & & & &\\
% 50k &	83.3/71.1 & 72.0/57.3 &	62.7/44.9 &	75.6/56.9 &	 73.4/57.5  \\
% 250k &	83.2/71.3 & 71.1/56.1 &	62.9/46.2 &	74.3/56.7 &	 72.9/57.6  \\
% \hline
% \end{tabular}
\caption{XQuAD (F1/EM) scores 
for different amounts of parallel data. Experiments were run with LargeWordOT. mBERT represents the case where we have no parallel datasets}
\label{table:diff_amount_parallel_results_xquad}
\vspace{-5mm}
\end{table}

\section{Shuffling different languages in one OT process}
\label{shuffling}
In all our experiments, we trained an independent OT per language pair. We additionally examined the impact of combining more than one non-English language in the same OT optimization versus learning independent OT per language. Hence, in each batch, we have pairs of sentences (non-English to their equivalents in English) drawn equally from all languages seen during training; remaining parameters are the same hence we back-propagate the loss values with the same number of computations. Combining sentences from different languages in one OT optimization leads to soft aligning all seen languages at once minimizing the cost of transferring knowledge from source to target. We observe consistent significant drop across languages in XNLI. The performance dropped for approximately 5.1\% for de, 3.8\% for es, 5.1\% for fr, and 9.1\% for bg. As we conflate sentences from different languages, the OT alignment optimization becomes harder especially that we follow batching strategy and languages can differ at different linguistic properties (e.g. syntactic structure ... etc). 

%answered in above paragraph @Sawsan:  How do we define source and target distribution when considering multiple languages per OT ?

\section{Examples}
\label{example_section}

\begin{table*}[h!]
\small
    \centering
    \addtolength{\tabcolsep}{-2.0 pt}

    \begin{tabular}{p{3cm} | p{3cm}} 
    \hline
\multicolumn{2}{c}{
\<
وأود في هذا الصدد أن أتناول بوجه خاص
قضية مدينة القدس ،
>   
} \\
\multicolumn{2}{c}{
\<
وهي قضية أساسية بالنسبة لجميع أعضاء المجموعة 
العربية
>   
} \\
\multicolumn{2}{c}{
In this regard , I wish to address in specific the issue of the City of Jerusalem ,}\\
\multicolumn{2}{c}{
a central issue for all of the Members of the Arab Group  }\\
\hline
\<وأود>&
I wish\\
\<في> & In,in\\
\<هذا> & this \\
\<الصدد> & regard\\
\<أن>  & to \\
\<أتناول> &  address\\ 
\<بوجه>  & 
\textbf{in the}\\
\<خاص> & specific \\
\<قضية>  & \textbf{ the,issue,of} \\
\<مدينة>  &\textbf{the} \\
\<القدس>
 & City of Jerusalem \\
\<قضية>  & issue \\
\<أساسية>  & Central\\ 
\<بالنسبة> & for,\textbf{of} \\
\<لجميع> & all,\textbf{ the} \\
\<أعضاء>&\textbf{ of}, Members,\textbf{the}  \\ 
\<المجموعة>
 & \textbf{Members},the,\textbf{Arab},Group \\
\<العربية>
 & Arab,\textbf{Group} \\
\\
\hline
\multicolumn{2}{c}{
\<
وبذلك اختتمت اللجنة مناقشتها العامة لهذا البند من جدول الأعمال .
>
}\\
\multicolumn{2}{c}{
The Committee thus concluded its general discussion on this agenda item .}
\\
\hline
\<وبذلك> & thus \\
\<اختتمت>   & concluded \\
\<اللجنة> &  Committee  \\
\<مناقشتها> &  its,\textbf{general}\\
\<العامة> &\textbf{ discussion}  \\
\<لهذا>  & this \\
\<البند>  &  agenda,item\\
\<من> & on \\
\<جدول> & \textbf{ The,this} \\
\<الأعمال>  & \textbf{ discussion},agenda\\
\\
\hline
\multicolumn{2}{c}{
\<
أربعة أعضاء من دول أوروبا الغربية ودول أخرى .
> }\\
\multicolumn{2}{c}{
Four members from Western European and other States .}\\
\hline
\<أربعة>&  Four\\
\<أعضاء>& members\\
\<من> & from \\
\<دول> 
&\textbf{Western} \\
\<أوروبا> & European \\
\<الغربية>&
\textbf{European,States}\\
\<ودول> & and,States \\ 
\<أخرى> & other \\
\hline
    \end{tabular}
    \caption{Alignment examples in Arabic. Words in \textbf{bold} are either errors or not direct alignment.}
    \label{examples_arabic}
\end{table*}

\begin{table*}[h!]
\small
    \centering
 \addtolength{\tabcolsep}{-2.0 pt}
    \begin{tabular}{p{3cm} | p{3cm}}     \hline
\multicolumn{2}{c}{
Zunächst wurde die für die Beitreibung der traditionellen Eigenmittel bzw.} \\
\multicolumn{2}{c}{
Zölle und 
Agrarausgleichsbeträge zu erhebende Prämie auf 25 Prozent erhöht} \\
\multicolumn{2}{c}{
First , the premium paid for the collection of traditional own resources , }\\
\multicolumn{2}{c}{
i.e. customs duty and agricultural levies , was increased to 25 \%}\\
\hline
Zunächst & First \\
wurde & , was \\
die & the \\
für & for \\
die & the \\
Beitreibung & premium,collection \\
der & of \\
traditionellen & collection,traditional,own,agricultural \\
Eigenmittel & resources \\
bzw. & i.e. \\
Zölle & customs,duty,agricultural,levies \\
und & and \\
Agrarausgleichsbeträge & duty,levies \\
zu & paid \\
erhebende & paid \\
Prämie & premium \\
auf & to \\
25 & 25 \\
Prozent & \% \\
erhöht & increased \\
\hline
\multicolumn{2}{c}{Mit anderen Worten : Die tschechische Rahmengesetzgebung in} \\
\multicolumn{2}{c}{diesem Bereich muß an die der Europäischen Union angepaßt und auch praktisch umgesetzt werden} \\
\multicolumn{2}{c}{
In other words , the Czech framework legislation in this area must be adapted and to all }\\
\multicolumn{2}{c}{all intents and purposes converted to that of the European Union}\\
\hline
Mit & In \\
anderen & other \\
Worten & words \\
: & , \\
Die & the \\
tschechische & Czech \\
Rahmengesetzgebung & framework,legislation \\
in & in \\
diesem & this \\
Bereich & area \\
muß & must \\
an & to,to,of \\
die & that,of \\
der & the \\
Europäischen & European \\
Union & Union \\
angepaßt & adapted,converted \\
und & and,and \\
auch & \textbf{all,intents,purposes} \\
praktisch & all,purposes \\
umgesetzt & adapted \\
werden & be \\
\hline
\multicolumn{2}{c}{Mit der Einführung des Euro ist das Wechselkursrisiko verschwunden} \\
\multicolumn{2}{c}{The exchange rate risk has disappeared with the advent of the euro} \\
\hline
Mit & with \\
der & the \\
Einführung & advent,of \\
des & the \\
Euro & euro \\
ist & has \\
das & The \\
Wechselkursrisiko & exchange,rate,risk \\
verschwunden & disappeared \\
\hline
    \end{tabular}
    \caption{Alignment examples in German. Words in \textbf{bold} are either errors or not direct alignment.}
    \label{german_examples}
\end{table*}

\begin{table*}[!h]
\small
\centering
\begin{tabular}{p{10cm} }
\hline
ar: \<الاتفاقية الإطارية>
\\
en: \textit{``FCCC / SBSTA / 2002 / L.23 / Add.1 Article 6 of the Convention''} \\
\hline
de: \textit{``Herr Präsident !''} \\
en: \textit{``Mr President , we are dealing here with sectors which have been excluded for a long time .''} \\
 \hline
de: \textit{``Im übrigen wurden die Abhängigkeitsverhältnisse eher verstärkt , als daß die Schuldenprobleme wirklich geklärt worden wären''} \\
en: \textit{``An analysis of the situation would seem to be more of a diagnosis as the details available and the same explanatory statement leave several signs of this imbalance the world is suffering''} \\
 \hline
\end{tabular}
\caption{Examples of incorrect pairs in parallel corpus.}
\label{table:bad_alignment_examples}
\vspace{-5mm}
\end{table*}

\end{document}